\documentclass[11pt,a4paper]{article}
\usepackage[margin=24mm,top=22mm,bottom=24mm,headheight=14pt]{geometry}
\usepackage[T1]{fontenc}
\usepackage[utf8]{inputenc}
\usepackage{lmodern,microtype,amsmath,amssymb,amsthm,booktabs,array,tabularx,graphicx}
\usepackage{tikz}
\usetikzlibrary{arrows.meta,positioning,calc}
\usepackage[font=small,labelfont=bf]{caption}
\usepackage[numbers,sort&compress]{natbib}
\usepackage{xcolor,fancyhdr,placeins}
\usepackage[hidelinks]{hyperref}
\hypersetup{pdftitle={Zero-Dimensional Attention Topology for Linguistic Acceptability},pdfauthor={Saeed AlMarri},pdfsubject={Linguistic acceptability and topological analysis of BERT attention},pdfkeywords={topological data analysis, linguistic acceptability, BERT attention, persistent homology}}
\pdftrailerid{}
\definecolor{ink}{HTML}{23384D}
\definecolor{panel}{HTML}{F2F5F8}
\definecolor{accent}{HTML}{356A8A}
\newtheorem{observation}{Observation}
\title{\vspace{-9mm}\LARGE\bfseries Zero-Dimensional Attention Topology\\for Linguistic Acceptability}
\author{Saeed AlMarri\\[3pt]
\small Department of Engineering Science and Management\\
\small Khalifa University of Science and Technology, Abu Dhabi, UAE\\
\small\texttt{100061460@ku.ac.ae}}
\date{14 November 2026}
\begin{document}
\maketitle
\thispagestyle{plain}
\begin{abstract}
Topological summaries of Transformer attention provide sentence representations for linguistic acceptability classification. We examine a zero-dimensional persistent-homology ($H_0$) representation that summarizes minimum-spanning-tree merge values from 144 symmetrized BERT attention graphs into 1,440 features. The retrospective evaluation uses 8,517 downstream training sentences and 521 in-domain and 497 out-of-domain CoLA validation sentences after data-integrity filtering. The recorded out-of-domain Matthews correlation coefficients are 0.438 for the topological representation and 0.508 for attention statistics, compared with 0.545 for BERT. Attention-feature results are retained aggregate measurements; BERT predictions are rescored and length controls are refitted from the accompanying records. Analytical results show that the representation retains token-count information and discards a class of directional attention differences. Source resampling of BERT predictions also demonstrates the sensitivity of uncertainty estimates to the sampling unit. The reported results do not demonstrate a predictive advantage for this fixed topological representation over simpler attention summaries. They identify representation properties and evaluation controls relevant to subsequent studies of attention topology.
\end{abstract}
\noindent\textbf{Keywords:} topological data analysis; linguistic acceptability; BERT attention; persistent homology.

\section{Introduction}
Linguistic acceptability classification provides a setting in which to investigate what language-model representations retain about sentence structure. The Corpus of Linguistic Acceptability (CoLA) contains acceptability judgments drawn from the linguistics literature \citep{warstadt2019}. Transformer attention represents directed, weighted token interactions that can be analyzed as graphs. Previous studies have used the topology of these graphs to improve acceptability classification and investigate linguistic phenomena \citep{cherniavskii2022,proskurina2023}.

The predictive value of a topological representation depends on the information retained by its construction and on the controls used to evaluate it. A graph summary may capture useful structure while also retaining sentence-length information or discarding directional interactions. Predictive performance and explanatory faithfulness therefore require separate assessment.

This study examines a fixed zero-dimensional persistent-homology ($H_0$) representation derived from a supervised BERT checkpoint \citep{devlin2019}. Each sentence is represented by summaries of the component-merge values of 144 symmetrized attention graphs. The central question is whether this representation demonstrates an advantage over simpler attention statistics or the BERT classifier on matched CoLA validation examples.

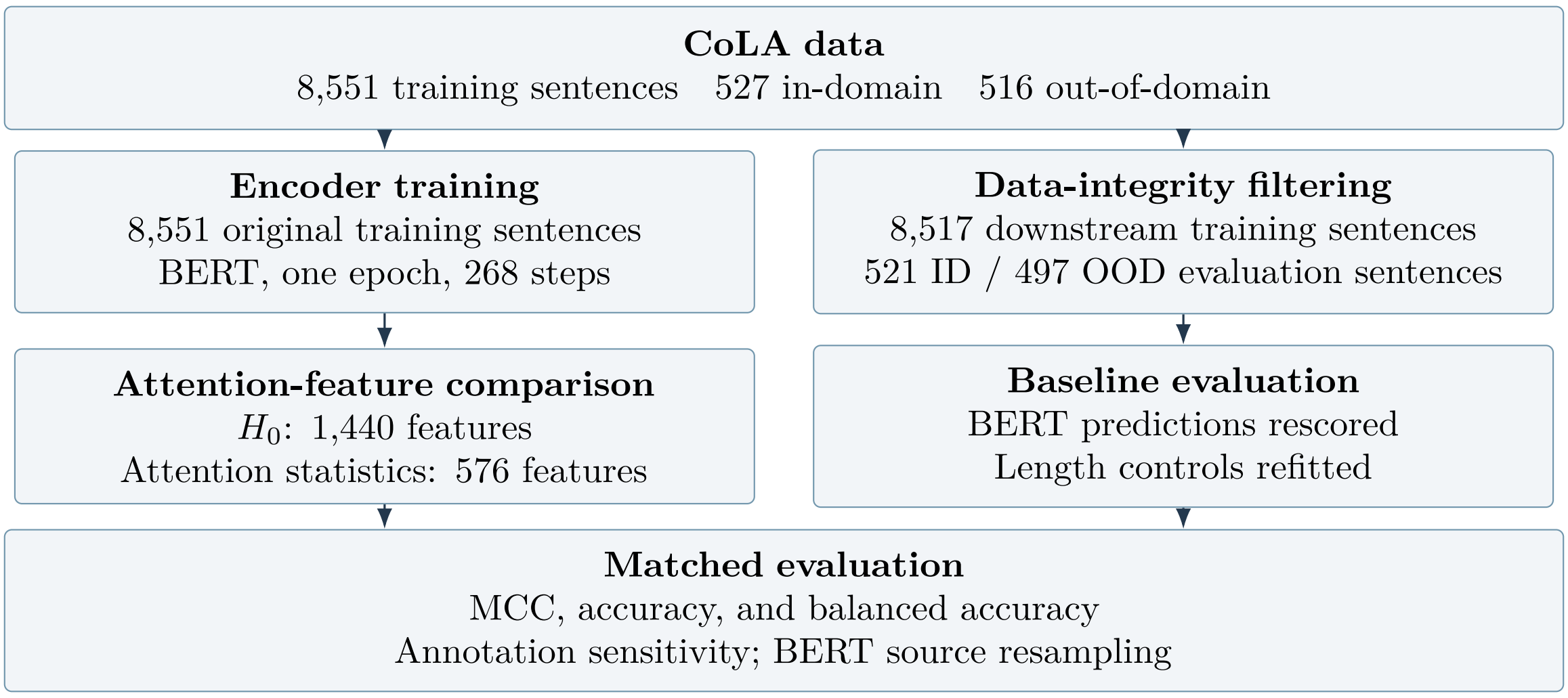
\begin{figure}[tbp]
\centering
\begin{tikzpicture}[font=\small,>=Latex,
 box/.style={draw=accent!70,fill=panel,rounded corners=2pt,align=center,inner sep=6pt,minimum height=12mm},
 arrow/.style={->,line width=.6pt,draw=ink}]
\node[box,text width=14.8cm] (data) at (0,0) {\textbf{CoLA data}\\8,551 training sentences\quad 527 in-domain\quad 516 out-of-domain};
\node[box,text width=6.8cm] (encoder) at (-3.9,-1.6) {\textbf{Encoder training}\\8,551 original training sentences\\BERT, one epoch, 268 steps};
\node[box,text width=6.8cm] (filter) at (3.9,-1.6) {\textbf{Data-integrity filtering}\\8,517 downstream training sentences\\521 ID / 497 OOD evaluation sentences};
\draw[arrow] (data.south -| encoder.north)--(encoder.north);
\draw[arrow] (data.south -| filter.north)--(filter.north);
\node[box,text width=6.8cm] (features) at (-3.9,-3.5) {\textbf{Attention-feature comparison}\\$H_0$: 1,440 features\\Attention statistics: 576 features};
\node[box,text width=6.8cm] (baselines) at (3.9,-3.5) {\textbf{Baseline evaluation}\\BERT predictions rescored\\Length controls refitted};
\draw[arrow] (encoder)--(features);
\draw[arrow] (filter)--(baselines);
\node[box,text width=14.8cm] (eval) at (0,-5.3) {\textbf{Matched evaluation}\\MCC, accuracy, and balanced accuracy\\Annotation sensitivity; BERT source resampling};
\draw[arrow] (features.south)--(features.south |- eval.north);
\draw[arrow] (baselines.south)--(baselines.south |- eval.north);
\end{tikzpicture}
\caption{Study design. The attention-feature readouts use the filtered downstream training set and are compared on the same evaluation sentences as the baselines. Their results are reported aggregates; BERT and length-control results are recalculated from the accompanying records. ID and OOD denote the in-domain and out-of-domain labeled validation partitions.}
\label{fig:design}
\end{figure}

The study combines a retrospective comparison with an analysis of the representation's mathematical properties. It documents the evaluation partitions and distinguishes encoder training exposure from downstream fitting data; establishes how the feature map retains token count and loses certain directional differences; and examines uncertainty under sentence and linguistic-source resampling. The recorded point estimates do not demonstrate an advantage for the tested $H_0$ representation. Figure~\ref{fig:design} summarizes the design.

\section{Related work}
\citet{cherniavskii2022} analyze attention topology for acceptability judgments across multiple languages and in minimal-pair evaluation. \citet{proskurina2023} study English CoLA and Russian RuCoLA using directed attention-graph features, including chordality and matching number, together with analyses of attention-head behavior. The present study examines a narrower $H_0$ construction. Differences in feature families and evaluation protocols preclude direct comparison of their published scores with the matched results reported here.

Persistent homology describes how topological structure changes across a filtration \citep{chazal2021}. Persistence images and landscapes provide vector summaries of persistence diagrams \citep{adams2017,bubenik2015}. Here, each sentence receives a vector of $H_0$ connectivity summaries computed from its attention graphs; the representation does not include higher-dimensional cycles.

Work on attention-based explanation distinguishes predictive information from faithful explanations of model behavior \citep{jain2019,wiegreffe2019}. Research on probing also motivates the use of controls when interpreting linguistic information in a representation \citep{hewitt2019}. The attention-statistics and length baselines in this study assess predictive utility for acceptability classification. Broader resources, such as MELA \citep{zhang2024}, provide multilingual evaluation settings beyond the English CoLA scope considered here.

\section{Data and experimental setup}
\subsection{Model and result provenance}
The saved model configuration identifies \texttt{bert-base-cased}, with 12 layers, 12 attention heads per layer, and hidden size 768. The trainer state records one epoch and 268 optimization steps. The recorded training configuration specifies a learning rate of $3\times10^{-5}$, weight decay 0.01, batch size 32, and class-balanced loss. The encoder was trained on the original 8,551 training rows; downstream feature-based classifiers use the filtered training set described below.

Results are distinguished by their provenance: \textbf{R} denotes reported aggregate measurements for the attention-feature models; \textbf{V} denotes reconstructed data counts or BERT scores recalculated from saved predictions; and \textbf{N} denotes refitted length controls or additional calculations. The $H_0$ and attention-statistics scores and paired intervals are retained as reported aggregates. The accompanying records support prediction-level recalculation of the BERT and length-control results.

\subsection{Partitions and duplicate filtering}
The source data contain 8,551 training rows, 527 in-domain rows, and 516 out-of-domain rows. Both evaluation partitions are labeled CoLA validation data. In particular, the source file named \texttt{test.csv} is the out-of-domain validation partition, not the hidden GLUE test set. The evaluation is retrospective.

A normalized sentence key is obtained by Unicode NFKC normalization, case folding, and whitespace collapsing. Within the training partition, all rows in a conflicting-label key group are excluded; for consistent-label repetitions, the first row is retained. A subsequent token-sequence check removes two additional consistent-label duplicates. Evaluation filtering excludes matches to all original 8,551 encoder-exposure rows, within-partition conflicting-label groups, and previously retained evaluation duplicates. In-domain data are processed before out-of-domain data. Token-equivalent exposure and evaluation repetitions are checked after the text-key filters.

The conflict policy is partition-local, so evaluation labels do not determine training exclusions. Training filtering removes 22 rows from 11 conflicting-label groups, 10 repeated normalized-text rows, and two token-equivalent rows, leaving 8,517 examples with 6,001 positive labels (Table~\ref{tab:flow}). Conflicting labels can reflect contextual or annotation differences; filtering does not adjudicate the underlying linguistic judgment.

\begin{table}[tbp]
\centering\small
\caption{Reconstructed sample flow (V). The strict subsets require exact normalized-text and label concordance with both annotation tables and are nested within the primary evaluation sets.}
\label{tab:flow}
\begin{tabular}{lrrr}
\toprule
 & Training & In-domain & Out-of-domain\\
\midrule
Original rows &8,551&527&516\\
Training-exposure matches removed &--&6&12\\
Conflicting-label rows removed &22&0&6\\
Repeated normalized-text rows removed &10&0&1\\
Additional token-equivalent rows removed &2&0&0\\
\midrule
Primary retained rows &8,517&521&497\\
Primary positive labels &6,001&361&339\\
Additional strict exclusions &--&3&2\\
Strict retained rows &8,517&518&495\\
\bottomrule
\end{tabular}
\end{table}

\subsection{Tokenization and annotation concordance}
Token sequences are reconstructed from the saved cased WordPiece configuration, retaining special tokens and using a maximum length of 64. The implementation follows the BERT tokenization procedure \citep{berttokenizer}. Across 9,594 rows, it identifies 43 repeated token-sequence groups, including 19 spanning partitions. Valid-token ranges are 4--47 for training, 5--31 in-domain, and 5--36 out-of-domain, so no reconstructed sequence requires truncation. Retained evaluation sequences are disjoint from the original encoder-exposure sequences under this reconstruction.

The annotation-concordant sensitivity subset retains a primary evaluation row only when both annotation tables contain the same normalized sentence with a single matching label. This excludes five additional rows: in-domain indices 53, 329, and 496, and out-of-domain indices 201 and 203, using zero-based source indices. Two exclusions concern label disagreements and three concern transcription differences. Labels and fitted models are unchanged for this sensitivity analysis.

\section{Representation and evaluation protocol}
\subsection{From attention to zero-dimensional persistence}
For each sentence and layer--head pair, let $A\in\mathbb{R}^{n\times n}$, with $n\ge2$, be the nonnegative attention matrix after removing padding rows and columns while retaining \texttt{[CLS]} and \texttt{[SEP]}. For positive row sums, define
\begin{equation}
 P_{ij}=\frac{A_{ij}}{\sum_{k=1}^{n}A_{ik}},\qquad
 D_{ij}=\begin{cases}
 1-\frac{P_{ij}+P_{ji}}{2},&i\ne j,\\
 0,&i=j.
 \end{cases}
 \label{eq:distance}
\end{equation}
The implementation uses float64 arithmetic and clips off-diagonal weights to $[0,1]$. The complete undirected graph includes edge $(i,j)$ at filtration value $t$ when $D_{ij}\le t$. All vertices are born at zero. The $n-1$ finite component-merge values equal the edge weights of a minimum spanning tree (MST); the final essential component is omitted \citep{chazal2021,kruskal1956}. This filtration does not require the triangle inequality.

Writing the finite deaths as $d_1,\ldots,d_{n-1}$, the per-head feature vector is
\begin{equation}
 \phi(D)=\bigl(\bar d,\sigma_d,Q_{.10},Q_{.25},Q_{.50},Q_{.75},Q_{.90},
                 d_{\max},S,H_d\bigr),
 \label{eq:features}
\end{equation}
where $\sigma_d$ is the population standard deviation, quantiles use linear interpolation, $S=\sum_r d_r$, and, for $S>0$,
\begin{equation}
 H_d=-\sum_{r:q_r>0}q_r\log q_r,\qquad q_r=d_r/S.
\end{equation}
The entropy is assigned zero when $S=0$. Concatenating all 144 heads yields 1,440 features. This representation retains merge-value distributions but not their labeled token endpoints.

\subsection{Controls and classifiers}
The attention-statistics control uses four quantities per head from the same normalized matrix $P$: mean row entropy, mean diagonal attention, mean maximum attention, and mean asymmetry,
\begin{equation}
 -\frac{1}{n}\sum_{i,j:P_{ij}>0}P_{ij}\log P_{ij},\quad
 \frac{\operatorname{tr}P}{n},\quad
 \frac{1}{n}\sum_i\max_jP_{ij},\quad
 \frac{1}{n^2}\sum_{i,j}|P_{ij}-P_{ji}|.
 \label{eq:stats}
\end{equation}
It has 576 features. The token-length control uses the number of valid tokens, including special tokens. The text-length control uses whitespace word count, Unicode character count, and mean word character length. The majority control predicts the most frequent retained training label. BERT predictions are read from the saved prediction files, whose sequential indices and prediction-writing code support alignment with the source rows.

All feature-based classifiers use training-only standardization followed by logistic regression with $C=1$, the \texttt{lbfgs} solver, a maximum of 3,000 iterations, no class weighting, and the default decision rule. The recorded protocol specifies no hyperparameter search or head selection, with convergence in 515 iterations for $H_0$ and 357 for attention statistics. The length controls use the same settings, including seed 42, and converge. Identical settings do not imply equal effective capacity or optimal regularization across representations of different dimensions.

\subsection{Metrics and uncertainty}
\label{sec:uncertainty}
Matthews correlation coefficient (MCC) is the primary metric, with accuracy and balanced accuracy as complementary summaries. MCC is assigned zero when its denominator is zero, including for all-positive predictions. The reported attention-feature comparison uses 2,000 paired sentence-bootstrap replicates with seed 42. Each retained normalized sentence is unique, so resampling sentence groups is equivalent to resampling retained rows.

For BERT and the length and majority controls, 95\% percentile intervals are recalculated from 2,000 resamples using \texttt{numpy.random.default\_rng(42)}, reinitialized separately for each partition and evaluation policy. A source-level diagnostic instead resamples linguistic source codes, retaining all rows from each selected code and drawing as many codes as the partition contains. Both procedures condition on fitted predictions and exclude variation from encoder retraining and model selection. Source resampling is treated as exploratory, particularly for the six-source out-of-domain partition.

\section{Properties of the representation}
\begin{observation}[Token count is recoverable]
Whenever $S>0$, the pair $(\bar d,S)$ in Equation~\eqref{eq:features} determines the number of valid tokens:
\begin{equation}
 n=1+\frac{S}{\bar d}.
 \label{eq:length}
\end{equation}
\end{observation}
\begin{proof}
There are $n-1$ finite deaths and $\bar d=S/(n-1)$. Dividing gives the result.
\end{proof}
Thus, the feature vector retains length information even when a length-only linear classifier performs poorly. Standardization with nonzero scales is an invertible affine transformation and does not remove this information. The identity does not imply that the fitted linear classifier uses length, because forming a ratio is nonlinear. For uniform attention, every death is $1-1/n$, $S=(n-1)^2/n$, and $H_d=\log(n-1)$.

\begin{observation}[Symmetrization removes directional circulation]
If $K^\top=-K$, $K\mathbf{1}=0$, and $P+K$ remains nonnegative, then $P$ and $P+K$ are row-stochastic and have the same $D$ in Equation~\eqref{eq:distance}. They therefore produce identical $H_0$ summaries.
\end{observation}
\begin{proof}
The zero row sums preserve stochasticity, and $(P+K)+(P+K)^\top=P+P^\top$.
\end{proof}
For a three-token example, take $P=\mathbf{1}\mathbf{1}^\top/3$ and
\begin{equation}
 K=\epsilon\begin{pmatrix}0&1&-1\\-1&0&1\\1&-1&0\end{pmatrix},
 \qquad 0<\epsilon<1/3.
\end{equation}
The directed circulation changes while all finite deaths remain $2/3$. The asymmetry statistic in Equation~\eqref{eq:stats} changes from zero to $4\epsilon/3$. The feature families therefore retain different information, although this example does not establish the cause of their empirical performance difference. Reindexing vertices of a fixed attention matrix likewise leaves the MST death multiset unchanged; rewriting or permuting a sentence can change the attention matrix itself.

The accompanying dense-Prim implementation agrees with an independent complete-edge Kruskal implementation on 140 synthetic float16-input matrices of sizes 2, 3, 5, 8, 16, 32, and 64, with zero maximum death-value discrepancy. Tests also cover vertex permutation, the length identity, uniform attention, zero-weight edges, directional circulation, and invalid inputs. These checks validate the reference implementation and the stated algebraic properties.

\section{Results}
\subsection{Classification performance}
\begin{figure}[tbp]
\centering\includegraphics[width=\linewidth]{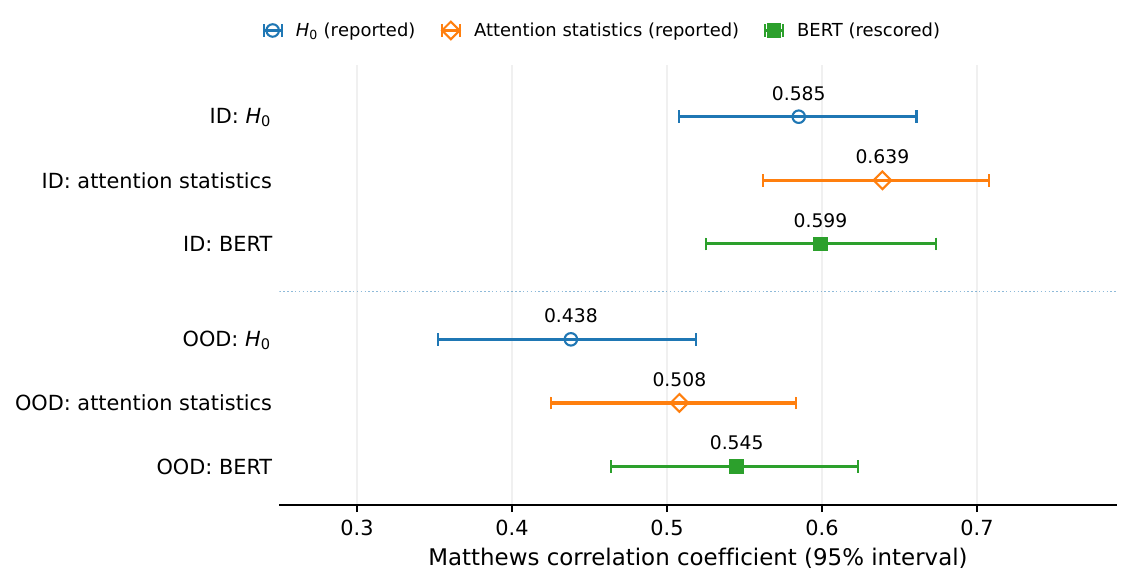}
\caption{MCC on matched primary evaluation sentences. Open markers denote reported attention-feature results; filled markers denote BERT scores recalculated from saved predictions. Error bars are marginal 95\% sentence-bootstrap intervals. Their overlap is not a paired test.}
\label{fig:mcc}
\end{figure}

Table~\ref{tab:primary} and Figure~\ref{fig:mcc} summarize the primary comparison. Recalculated BERT MCC is 0.599 in-domain and 0.545 out-of-domain. Both length models predict the positive class throughout both evaluation sets, matching the majority control. Their accuracy of approximately 0.68--0.69 reflects the class proportions rather than discrimination between acceptable and unacceptable sentences.

The reported $H_0$ point estimates are lower than those of attention statistics in both partitions. BERT is below attention statistics in-domain but above both attention-feature classifiers out-of-domain.

\begin{table}[tbp]
\centering\small
\caption{Primary evaluation. R denotes reported attention-feature aggregates, V denotes BERT rescored from saved predictions, and N denotes refitted length controls or the recomputed majority control. BERT and control intervals use the recalculated sentence bootstrap.}
\label{tab:primary}
\begin{tabular}{llcrrr}
\toprule
Partition & Model & Status & MCC [95\% interval] & Accuracy & Bal. acc.\\
\midrule
ID ($n=521$)&$H_0$&R&0.585 [0.508, 0.661]&0.829&0.779\\
&Attention statistics&R&0.639 [0.562, 0.708]&0.852&0.794\\
&BERT&V&0.599 [0.525, 0.674]&0.835&0.785\\
&Token length&N&0.000 [0.000, 0.000]&0.693&0.500\\
&Text length&N&0.000 [0.000, 0.000]&0.693&0.500\\
&Majority&N&0.000 [0.000, 0.000]&0.693&0.500\\
\midrule
OOD ($n=497$)&$H_0$&R&0.438 [0.352, 0.519]&0.765&0.709\\
&Attention statistics&R&0.508 [0.425, 0.583]&0.795&0.740\\
&BERT&V&0.545 [0.464, 0.623]&0.811&0.753\\
&Token length&N&0.000 [0.000, 0.000]&0.682&0.500\\
&Text length&N&0.000 [0.000, 0.000]&0.682&0.500\\
&Majority&N&0.000 [0.000, 0.000]&0.682&0.500\\
\bottomrule
\end{tabular}
\end{table}

\subsection{Paired contrasts and annotation sensitivity}
\begin{figure}[tbp]
\centering\includegraphics[width=\linewidth]{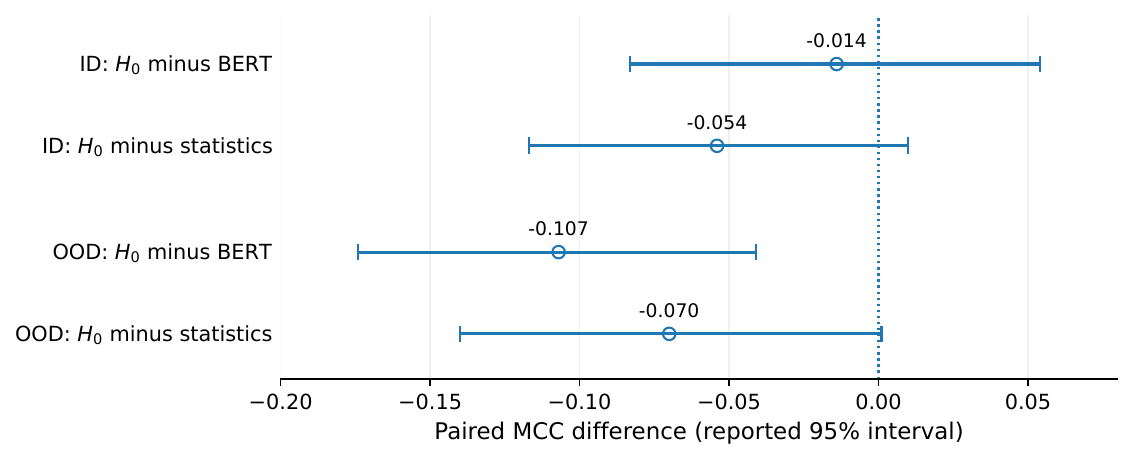}
\caption{Reported paired MCC contrasts (R). Negative values favor the comparator. The out-of-domain BERT contrast excludes zero under the recorded sentence-bootstrap procedure; the out-of-domain attention-statistics contrast narrowly includes zero.}
\label{fig:contrasts}
\end{figure}

The reported out-of-domain $H_0$-minus-BERT MCC difference is $-0.107$, with a paired interval of $[-0.174,-0.041]$ (Figure~\ref{fig:contrasts}). The difference relative to attention statistics is $-0.070$, with an interval of $[-0.140,0.001]$. Both in-domain contrast intervals include zero. These descriptive intervals are conditional on the recorded predictions and are unadjusted for multiple comparisons; an interval containing zero does not establish equivalence.

The annotation-concordant subsets preserve the point-estimate ordering (Table~\ref{tab:strict}). The reported strict out-of-domain $H_0$-minus-BERT interval is $[-0.181,-0.047]$; the corresponding attention-statistics contrast is $[-0.148,-0.002]$. The strict subsets reuse the same fitted models and differ from the primary evaluation by five sentences, so their results are sensitivity checks rather than independent confirmation.

\begin{table}[tbp]
\centering\small
\caption{Annotation-concordant sensitivity analysis. Values are MCC. R and V have the same meanings as in Table~\ref{tab:primary}. Both length controls and the majority control remain at zero.}
\label{tab:strict}
\begin{tabular}{lrrrr}
\toprule
Partition & $n$ & $H_0$ (R) & Attention statistics (R) & BERT (V)\\
\midrule
In-domain &518&0.588&0.642&0.602\\
Out-of-domain &495&0.435&0.511&0.548\\
\bottomrule
\end{tabular}
\end{table}

\subsection{Variation across linguistic sources}
\begin{figure}[tbp]
\centering\includegraphics[width=\linewidth]{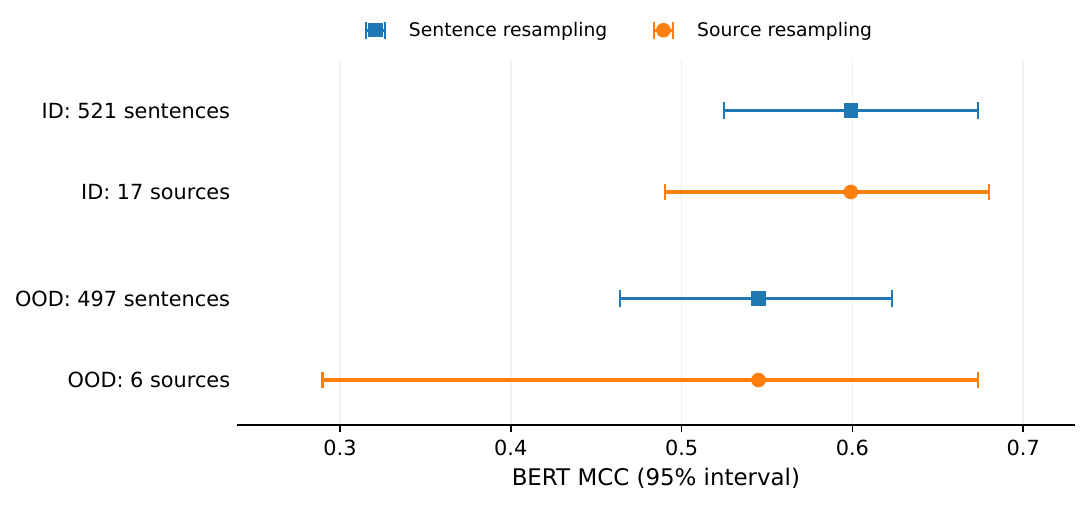}
\caption{Sensitivity of BERT uncertainty estimates to the resampling unit (N). Each interval uses 2,000 percentile-bootstrap replicates. Source resampling draws source codes with replacement and retains all sentences from each selected code, producing a variable sample size.}
\label{fig:sources}
\end{figure}

Out-of-domain source-specific BERT MCC ranges from 0.104 to 0.711, and the largest source contributes 210 of the 497 retained sentences. At the overall MCC of 0.545, the interval is $[0.464,0.623]$ under sentence resampling and $[0.290,0.673]$ under source resampling (Figure~\ref{fig:sources}). The corresponding in-domain intervals are $[0.525,0.674]$ and $[0.491,0.680]$.

The wider out-of-domain source interval illustrates sensitivity to the resampling unit. With only six source groups, this is an exploratory diagnostic of BERT predictions. It does not estimate a source-level contrast between the attention-feature classifiers.

\FloatBarrier
\section{Discussion and limitations}
The recorded comparison does not demonstrate a predictive advantage for the tested symmetrized $H_0$ representation. Its MCC exceeds the length and majority controls but falls below the attention-statistics point estimates in both partitions and below BERT out-of-domain. These results support the use of simple attention summaries as comparators when assessing topological sentence representations.

The analytical properties help characterize the representation. Poor length-only prediction does not establish the absence of length information in the topological features, since token count is recoverable from two coordinates. Symmetrization also discards directional differences detected by the attention-statistics control. Neither property establishes the cause of the observed performance gap. Dimensionality, regularization, the fixed linear classifier, and complementary information warrant matched ablation studies.

The empirical scope is one English dataset, one supervised BERT checkpoint trained for one epoch, and one recorded fit per attention representation. Other encoders, training seeds, directed topology, higher-dimensional homology, filtrations, and combined representations are outside this comparison. The evaluation uses previously available labeled validation partitions, so earlier inspection or model selection cannot be excluded. Duplicate filtering does not remove prior encoder exposure or all shared linguistic constructions. Token-equivalence checks depend on the reconstructed tokenizer and do not establish bitwise equivalence with the original runtime.

The $H_0$ and attention-statistics feature matrices, fitted classifiers, and per-sentence predictions are not part of the available evaluation record. Their scores and paired intervals therefore remain reported aggregates rather than independently regenerated prediction-level results. Bootstrap intervals also condition on fitted predictions and omit retraining and selection variability. The source diagnostic illustrates one component of this uncertainty. Conclusions concern predictive performance under the stated protocol; causal interpretability and computational efficiency were not evaluated.

\section{Conclusion}
This retrospective CoLA comparison finds no demonstrated advantage for a fixed $H_0$ summary of symmetrized BERT attention over simpler attention statistics. The representation retains token-count information while discarding a class of directional interactions, and uncertainty estimates for BERT vary with the resampling unit. Together, these findings motivate explicit analysis of representation invariances, attention-statistics controls, and source dependence in future evaluations of attention topology.

\section*{Data and code availability}
The accompanying \texttt{anc} directory contains 9,594 text-free audit records, 1,043 original-partition BERT prediction records with evaluation masks, length-control parameters, metric and source-level summaries, reported attention-feature aggregates, input hashes, and executable reconstruction and verification scripts. It also includes the tested $H_0$ reference implementation and figure-generation code. These materials support reconstruction of filtering decisions and recalculation of the baseline and sensitivity analyses without the original checkpoint. Rebuilding tokenization-derived records from raw text additionally requires the inputs identified in the manifest. Raw sentence text, vocabulary, model weights, and attention arrays are not included. Dependencies and execution commands are documented in \texttt{anc/README.md}.

\FloatBarrier
\begingroup\small\setlength{\bibsep}{3pt}
\bibliographystyle{plainnat}
\bibliography{references}
\endgroup
\par
\appendix
\section{Supplementary evaluation results}
Tables~\ref{tab:confusion} and \ref{tab:source} report confusion counts and out-of-domain source metrics from the same saved BERT predictions. Bootstrap intervals follow the procedure in Section~\ref{sec:uncertainty}; the ancillary records also retain the previously reported intervals.

\begin{table}[!htbp]
\centering\small
\caption{BERT confusion counts. The positive class denotes acceptable sentences.}
\label{tab:confusion}
\begin{tabular}{llrrrrr}
\toprule
Partition & Policy & $n$ & TN & FP & FN & TP\\
\midrule
In-domain & Primary &521&105&55&31&330\\
In-domain & Strict &518&105&54&31&328\\
Out-of-domain & Primary &497&94&64&30&309\\
Out-of-domain & Strict &495&94&63&30&308\\
\bottomrule
\end{tabular}
\medskip
\caption{BERT performance on the six retained out-of-domain source groups.}
\label{tab:source}
\begin{tabular}{lrrr}
\toprule
Source code & Retained sentences & MCC & Accuracy\\
\midrule
\texttt{c-05} &32&0.663&0.844\\
\texttt{clc95} &76&0.104&0.645\\
\texttt{j\_71} &42&0.291&0.690\\
\texttt{s\_97} &55&0.607&0.818\\
\texttt{swb04} &210&0.711&0.881\\
\texttt{w\_80} &82&0.565&0.829\\
\bottomrule
\end{tabular}
\end{table}
\end{document}